\documentclass{article} 
\usepackage{iclr2027_conference,times}

\usepackage{amsmath,amsfonts,bm}

\def\eqref#1{equation~\ref{#1}}

\def\1{\bm{1}}

\DeclareMathAlphabet{\mathsfit}{\encodingdefault}{\sfdefault}{m}{sl}
\SetMathAlphabet{\mathsfit}{bold}{\encodingdefault}{\sfdefault}{bx}{n}

\usepackage{hyperref}
\usepackage{url}
\usepackage{hyperref}
\usepackage{url}
\usepackage{graphicx}
\usepackage{booktabs}
\usepackage{multirow}
\usepackage{array}
\usepackage{makecell}
\usepackage[table]{xcolor}
\usepackage{pifont}
\usepackage{subcaption}
\usepackage{algorithm}
\usepackage{algorithmic}
\usepackage{mathtools}
\usepackage{amsthm}
\usepackage{amssymb}
\usepackage{wrapfig}
\usepackage{tcolorbox}          
\usepackage{pifont}
\usepackage{enumitem}

\title{Outcome-Guided On-Policy Self-Distillation}

\author{
    ZheXu Wang\textmd{\textsuperscript{1,2}}\textmd{,}
   ~Mao-Lin Luo\textmd{\textsuperscript{1,2}}\textmd{,}
   ~Yankun Hong\textmd{\textsuperscript{3}}\textmd{,}~Zi-Hao Zhou\textmd{\textsuperscript{1,2}}\textmd{,} ~Bo Ye\textmd{\textsuperscript{1,2,4}}\textmd{,} ~Jian Zhao\textmd{\textsuperscript{4,5}}\textmd{,} \\~\textbf{Xialiang Tong}\textmd{\textsuperscript{3}}\textmd{,} ~\textbf{Min-Ling Zhang}\textmd{\textsuperscript{1,2}}\textmd{,} ~\textbf{Tong Wei}\textmd{\textsuperscript{1,2}}\footnotemark[2] \\
  \textsuperscript{1}School of Computer Science and Engineering, Southeast University, Nanjing 210096, China\\ \textsuperscript{2}Key Laboratory of Computer Network and Information Integration (Southeast University), \\
  ~~Ministry of Education, China\\ 
  \textsuperscript{3}Huawei Noah's Ark Lab\\ 
  \textsuperscript{4}Zhongguancun Academy \\
  \textsuperscript{5}Zhongguancun Institute of Artificial Intelligence \\
}

\begin{document}

\maketitle
\begin{abstract}
On-policy self-distillation (OPSD) provides denser token-level supervision and better computational efficiency than Reinforcement Learning with Verifiable Rewards (RLVR). However, this denser supervision may introduce substantial noise and training instability. Existing improvements often rely on high-variance per-token statistics and introduce extra hyperparameters and trade-offs. Based on the advantage formulation in RLVR, we analyze the OPSD objective from the same perspective, incorporating outcome correctness signals. We find that vanilla OPSD imposes insufficient penalties and excessive rewards on incorrect trajectories because it applies a fixed divergence objective regardless of outcome correctness. Furthermore, the reliability of teacher supervision is associated with both trajectory outcome and the cumulative average teacher entropy along the rollout. Based on these observations, we propose Outcome-Guided On-Policy Self-Distillation (OG-OPSD), which dynamically adapts both the divergence objective and distillation position according to binary outcome rewards and the cumulative average teacher entropy. Extensive experiments show that OG-OPSD consistently improves the performance of vanilla OPSD and multiple strong baselines in mathematical reasoning, multimodal reasoning, and out-of-distribution tasks across Qwen3 models at 1.7B, 4B, and 8B scales, as well as Qwen3-VL-2B.

\end{abstract}

\section{Introduction}

Reinforcement learning with verifiable rewards (RLVR) and on-policy distillation (OPD) have emerged as effective approaches to post-training large language models (LLMs). RLVR improves multi-step reasoning by optimizing automatically verifiable outcomes, avoiding costly human preference annotations \citep{guo2025deepseek,shao2024deepseekmath,jaech2024openai}. However, its commonly used binary outcome reward provides only a single signal for an entire trajectory, leaving token-level credit assignment unresolved. For example, outcome-supervised GRPO assigns the same group-normalized advantage to every token within a response \citep{shao2024deepseekmath}. OPD provides denser supervision by matching the teacher's next-token distributions along the student's own rollouts \citep{gu2024minillm,agarwal2024policy,lu2025onpolicydistillation}, but conventional OPD typically relies on a stronger external teacher.

On-policy self-distillation (OPSD) removes the need for a separate teacher by using the same model under different conditioning contexts \citep{zhao2026selfdistilled,shenfeld2026selfdistillation,hubotter2026reinforcement}. The student observes only the question, whereas the teacher additionally receives privileged information, such as a reference solution. This asymmetry enables dense self-supervision, but it does not guarantee that the resulting guidance is useful at every position. Recent studies report optimization instability and degraded reasoning performance in OPSD \citep{zhu2026many,kim2026why}, highlighting the need to determine how and where teacher supervision should be applied.

The distillation objective directly shapes this supervision. Forward and reverse Kullback-Leibler (KL) divergences weight teacher-student discrepancies differently, inducing distinct token-level learning signals. Vanilla OPSD applies a fixed divergence across all trajectories and positions. Recent work has explored adaptive objectives \citep{jin2026entropyaware,jung2025todi,jia2026asymmetric,wu2025rethinking}, as well as token reweighting and selection \citep{liu2026teacher,wang2026not,zhang2026tailoring,ziheng2026less}. Token probabilities and entropy characterize local model behavior, whereas outcome correctness provides a direct signal of task success. Although this outcome signal does not identify the contribution of each token, it can inform how dense supervision is used. Our central idea is to condition both the choice of divergence and the position of distillation under the guidance of outcome correctness.

\begin{figure}[!t]
\centering
\begin{subfigure}[b]{0.408\textwidth}
    \centering
    \includegraphics[width=\linewidth]{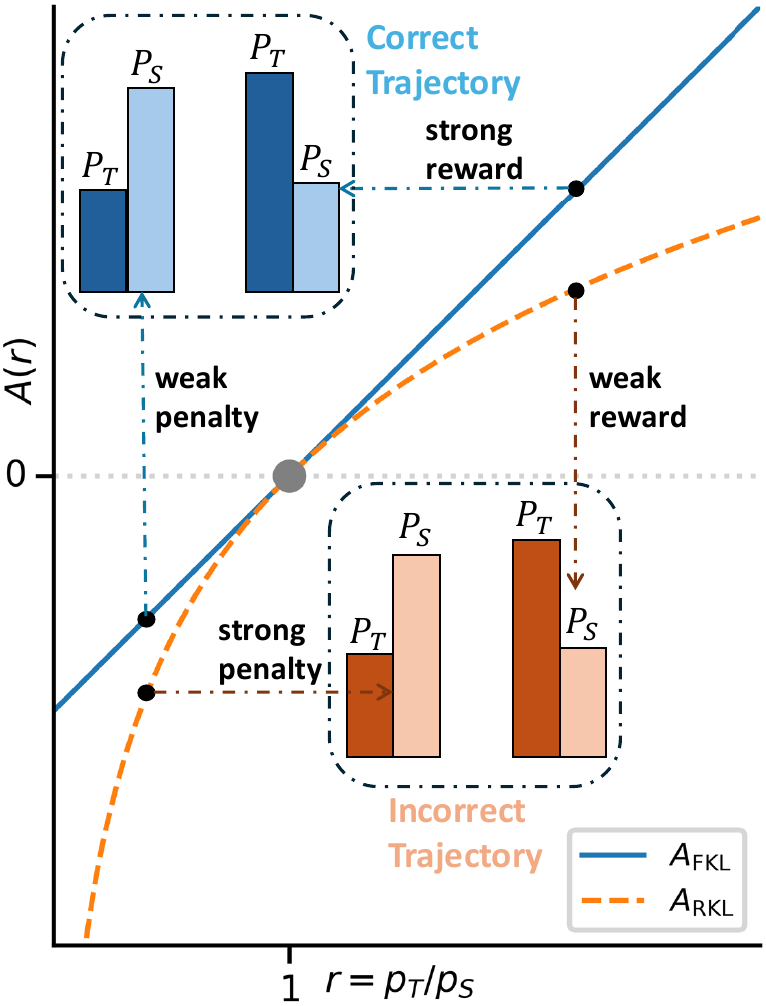}
    \caption{Outcome-guided divergence selection}
    \label{fig:introa}
\end{subfigure}
\hfill
\begin{minipage}[b]{0.29\textwidth}
    \begin{subfigure}[b]{\linewidth}
        \centering
        \includegraphics[width=\linewidth]{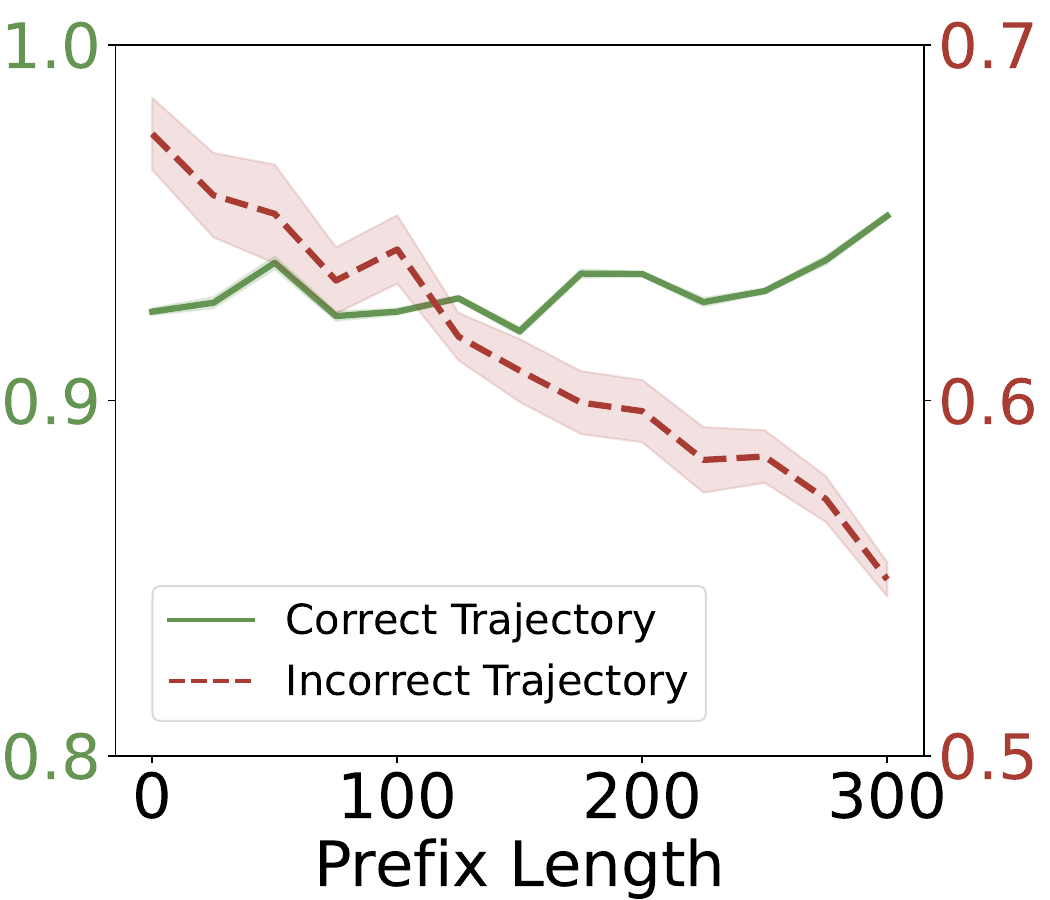}
        \caption{Qwen3-4B}
        \label{fig:introb}
    \end{subfigure}
    \vspace{0em}
    \begin{subfigure}[b]{\linewidth}
        \centering
        \includegraphics[width=\linewidth]{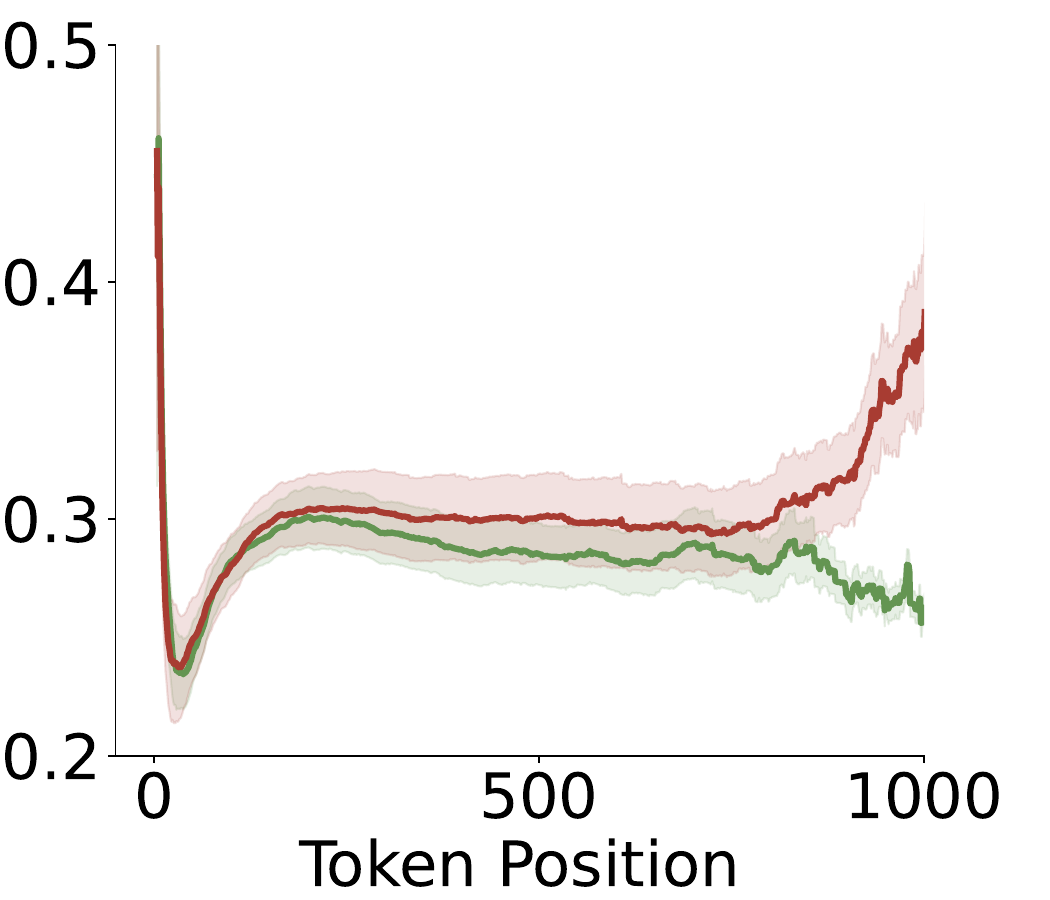}
        \caption{Qwen3-4B}
        \label{fig:introc}
    \end{subfigure}
\end{minipage}
\begin{minipage}[b]{0.29\textwidth}
    \begin{subfigure}[b]{\linewidth}
        \centering
        \includegraphics[width=\linewidth]{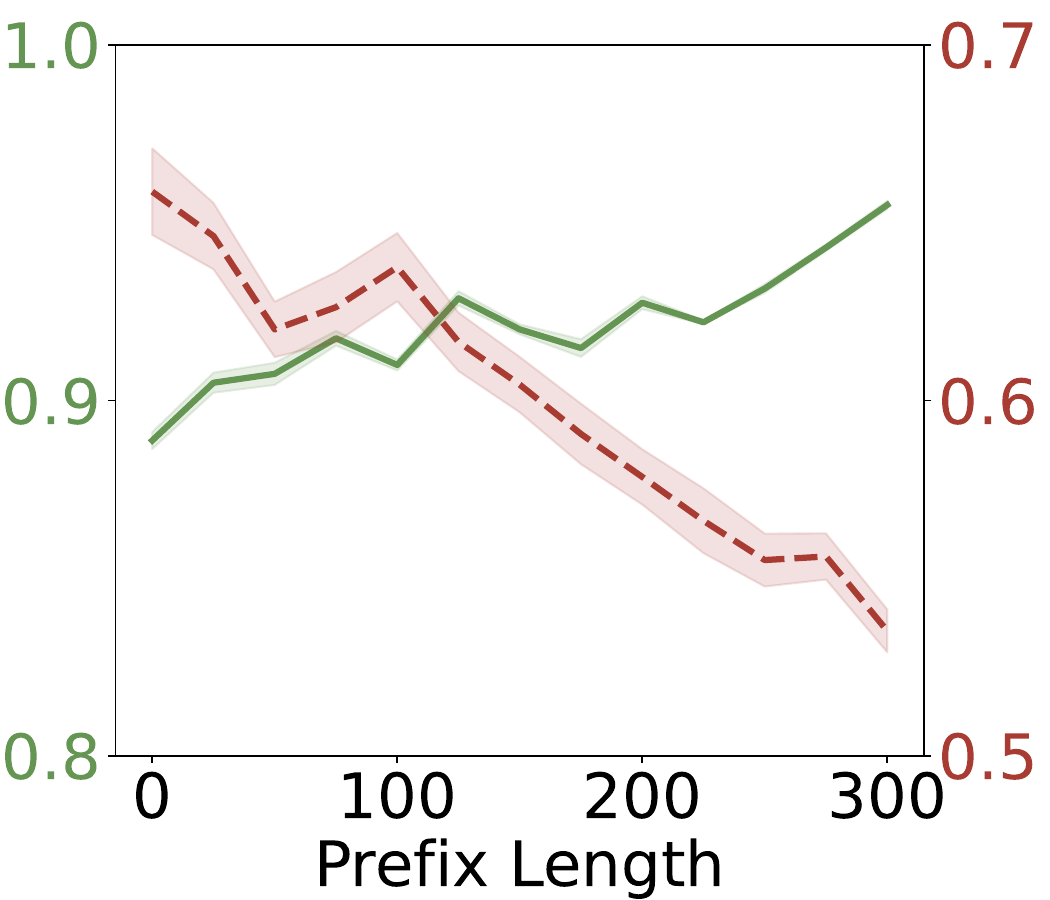}
        \caption{Qwen3-8B}
        \label{fig:introd}
    \end{subfigure}
    \vspace{0em}
    \begin{subfigure}[b]{\linewidth}
        \centering
        \includegraphics[width=\linewidth]{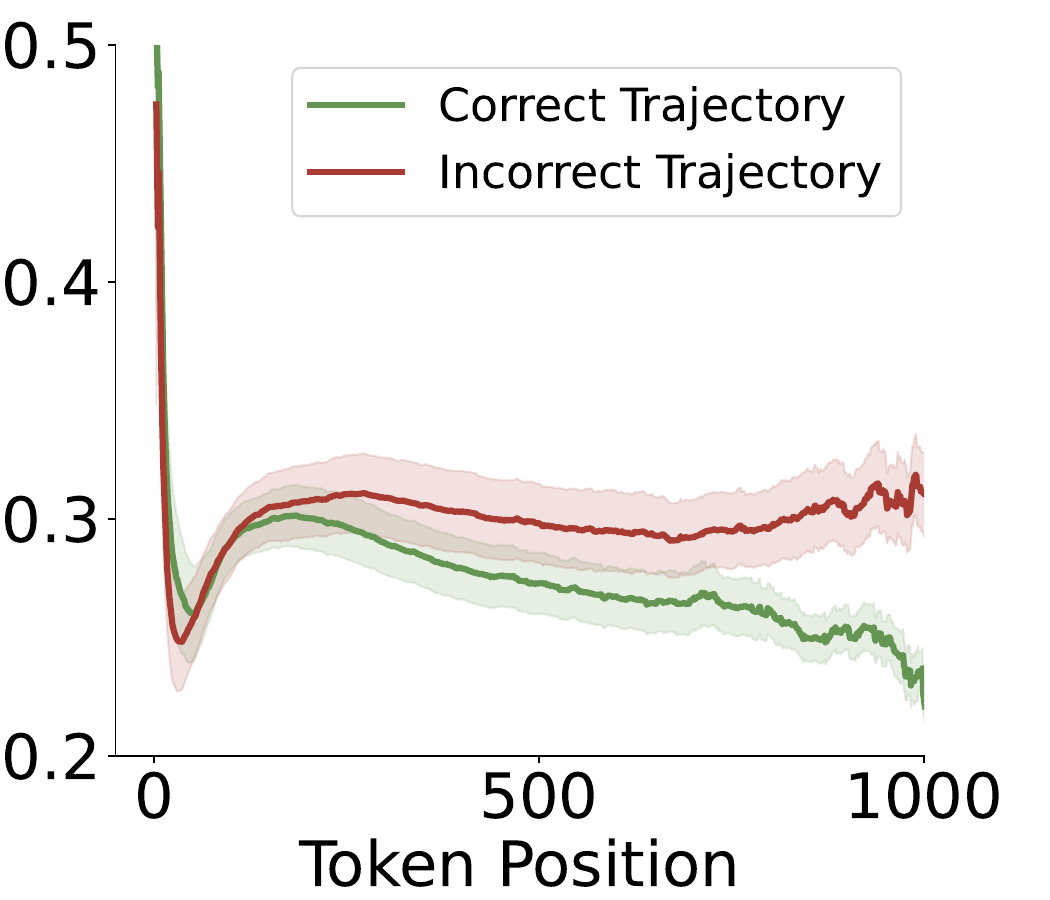}
        \caption{Qwen3-8B}
        \label{fig:introe}
    \end{subfigure}
\end{minipage}
\caption{
\textbf{(a)} Implicit token-level advantages under forward and reverse KL. $P_T$ and $P_S$ denote the probabilities assigned to a candidate token by the teacher and student, respectively; $A(\cdot)$ denotes the corresponding advantage function.
\textbf{(b,d)} Teacher completion accuracy (TCA) after resuming generation from student-generated reasoning prefixes of increasing length. Results are averaged over OpenThoughts-Math30K, DeepScaleR, MATH-500, and DeepMath datasets.
\textbf{(c,e)} Cumulative average teacher entropy along correct and incorrect student trajectories.
}
\label{fig:intro}
\end{figure}

We first analyze forward KL (FKL) and reverse KL (RKL) through their implicit token-level advantages (see their conceptual comparisons in Figure~\ref{fig:introa}). This formulation reveals an asymmetry in their weighting: FKL assigns stronger rewards to tokens underweighted by the student relative to the teacher and bounds penalties, whereas RKL imposes stronger penalties on tokens overestimated by the student. These properties motivate FKL on correct trajectories to learn teacher-preferred continuations while limiting negative feedback on successful behavior, and RKL on incorrect trajectories to reduce student overestimation relative to the teacher.

However, divergence selection alone cannot determine whether the teacher maintains reliability during deployment. We investigate this question by truncating student trajectories at varying lengths and asking the teacher to complete each prefix. Across model scales and mathematical reasoning benchmarks, teacher completion accuracy declines as prefixes from incorrect trajectories grow longer, while remaining comparatively stable on correct trajectories (see empirical results in Figures~\ref{fig:introb} and~\ref{fig:introd}). This pattern of prefix failure motivates restricting distillation on incorrect trajectories to shorter prefixes. We further examine the cumulative average teacher entropy, whose curves for correct and incorrect trajectories provide an empirical basis for estimating the truncation cutoff (see empirical results in Figures~\ref{fig:introc} and~\ref{fig:introe}).
We observe the same trend on Qwen3-1.7B (see empirical results in Figures~\ref{tca1b} and~\ref{tcae1b}).

Building on these findings, we propose \emph{Outcome-Guided On-Policy Self-Distillation} (OG-OPSD), which combines outcome-guided divergence selection with selective prefix truncation. OG-OPSD applies FKL throughout correct trajectories and restricts RKL to prefixes of incorrect trajectories. The cutoff is estimated from the last crossing of the cumulative average teacher entropy curves for the two outcome groups. Trajectory-level correctness thus determines how and where to distill, while the teacher supplies dense token-level supervision within the selected prefix.

Our contributions are summarized as follows:
\begin{itemize}
    \item We connect the asymmetric token-level weighting of FKL and RKL to trajectory correctness, and identify an outcome-dependent decline in teacher continuation accuracy as incorrect prefixes increase in length.
    \item We propose a simple but effective method called OG-OPSD, which combines outcome-guided KL divergence selection with an entropy-based prefix cutoff to control how and where self-distillation is applied.
    \item Experiments show consistent improvements over vanilla OPSD and strong baselines with Qwen3 models at different scales. Evaluations on Qwen3-VL and out-of-distribution datasets further support the effectiveness of the proposed method.
\end{itemize}

\section{Related Work}
\textbf{On-policy distillation.}
OPD combines student-generated rollouts with dense teacher supervision and has been adopted in recent LLM post-training pipelines \citep{xu2026deepseek,zeng2026glm,team2026kimi,yang2025qwen3}. OPSD removes the need for an external teacher by conditioning the same model on privileged information \citep{zhao2026selfdistilled,shenfeld2026selfdistillation,hubotter2026reinforcement}. Recent works improve the allocation of dense supervision through token weighting, selection, and rollout truncation. \citet{liu2026teacher} analyze how teacher reliability varies with token position and introduce position-dependent loss weights. \citet{wang2026not} select token positions according to teachability, distinguishing learnable teacher-student disagreement from incompatible predictions. \citet{zhang2026tailoring} use student entropy to determine the direction of supervision, encouraging agreement with the teacher at low-entropy positions and divergence at high-entropy positions to preserve exploration. Other studies address declining teacher reliability along student-generated prefixes. ESR restricts rollout length \citep{ziheng2026less}; Prune-OPD monitors teacher-student compatibility and accumulates drift signals to downweight unreliable supervision and truncate rollouts \citep{yang2026prune}; and IW-OPD weights token losses using cumulative prefix discrepancies \citep{xie2026position}. OG-OPSD makes the distillation span conditional on trajectory outcome: it retains full supervision on correct trajectories and uses cumulative average teacher entropy to estimate a prefix cutoff for incorrect trajectories.

\textbf{Divergence selection.}
The choice of divergence determines how teacher-student discrepancies translate into learning signals. MiniLLM minimizes sequence-level RKL \citep{gu2024minillm}, GKD supports alternative divergence objectives \citep{agarwal2024policy}, and DistiLLM introduces skewed KL divergences \citep{ko2024distillm}. Subsequent work adapts the objective to local distributional properties. EOPD augments RKL with FKL at positions with high teacher entropy \citep{jin2026entropyaware}. ToDi combines FKL and RKL using token-wise teacher-student log-probability ratios \citep{jung2025todi}, while AOPD preserves positive-advantage reinforcement and applies FKL guidance in non-positive-advantage regions \citep{jia2026asymmetric}. AKL adjusts the relative weights of FKL and RKL according to discrepancies in the head and tail of the next-token probability distribution \citep{wu2025rethinking}. Guided by an analysis of the implicit token-level advantages of FKL and RKL, OG-OPSD adopts a trajectory-level selection rule: FKL for correct rollouts and RKL for incorrect ones. This choice depends only on the binary verifier outcome and is coupled with selective prefix truncation to account for changes in teacher reliability.
\section{Method}
\label{sec:method}

OG-OPSD uses the binary outcome of the verifier to select the divergence and determine which token positions receive supervision. It applies forward KL to complete correct trajectories, and reverse KL to prefixes of incorrect trajectories, with the prefix length estimated from teacher entropy.

\subsection{Preliminaries}
\label{3.1}

OPSD uses the same model $\pi_\theta$ as both the teacher and the student, with different conditioning contexts \citep{zhao2026selfdistilled}. Given a question $x$, the student generates a response $y=(y_1,\ldots,y_N)\sim\pi_\theta(\cdot\mid x)$. The teacher also receives privileged information $y^\star$. At position $n$, let $s_n=(x,y_{<n})$ denote the question and the student-generated prefix. The student and the teacher define next-token distributions in the vocabulary $\mathcal{V}$:
\begin{equation}
p_S^n(\cdot) = \pi_\theta(\cdot\mid s_n), \qquad
p_T^n(\cdot) = \operatorname{sg}\!\left[\pi_\theta(\cdot\mid s_n,y^\star)\right],
\label{eq:opsd-policies}
\end{equation}
where $\operatorname{sg}$ denotes the stop-gradient. During optimization, sampled prefixes are treated as fixed, and gradients flow only through the student branch.

Vanilla OPSD averages local distillation losses over the entire response:
\begin{equation}
\mathcal{L}_{\mathrm{OPSD}}(\theta;y)
=\frac{1}{N}\sum_{n=1}^{N}d_n,
\label{eq:opsd-loss}
\end{equation}
using the same divergence at every position. We consider two forms of KL divergence:
\begin{align}
d_n^{\mathrm{FKL}}
&=D_{\mathrm{KL}}(p_T^n\Vert p_S^n)
=\sum_{v\in\mathcal{V}}p_T^n(v)\log\frac{p_T^n(v)}{p_S^n(v)},
\label{eq:local-fkl}\\
d_n^{\mathrm{RKL}}
&=D_{\mathrm{KL}}(p_S^n\Vert p_T^n)
=\sum_{v\in\mathcal{V}}p_S^n(v)\log\frac{p_S^n(v)}{p_T^n(v)}.
\label{eq:local-rkl}
\end{align}
OPSD provides supervision signals through the distribution differences of the teacher-student model, while RLVR provides supervision through a deterministic outcome reward $R(y)$ that can be automatically verified against ground truth. For a fixed question $x$, the expected-reward objective is $\mathcal{J}_{\mathrm{RLVR}}(\theta)=\mathbb{E}_{y\sim\pi_\theta(\cdot\mid x)}[R(y)]$, with REINFORCE gradient
\begin{equation}
\nabla_\theta\mathcal{J}_{\mathrm{RLVR}}(\theta)
=
\mathbb{E}_{y\sim\pi_\theta(\cdot\mid x)}
\left[
R(y)\sum_{n=1}^{N}\nabla_\theta\log p_S^n(y_n)
\right].
\label{RLVR}
\end{equation}
Here, the same trajectory-level reward weights every sampled token. Distillation provides finer-grained supervision through teacher-student discrepancies.

\subsection{Outcome-Guided Divergence Selection}
\label{3.2}

To compare how FKL and RKL allocate supervision, we express their local gradients in the score-function form of Eq.~\ref{RLVR}. For a fixed prefix, define $r_n(v)=p_T^n(v)/p_S^n(v)$. With gradients stopped through the teacher, the descent directions are
\begin{align}
-\nabla_\theta d_n^{\mathrm{FKL}}
&=\sum_{v\in\mathcal{V}}p_S^n(v)
\bigl(r_n(v)-1\bigr)\nabla_\theta\log p_S^n(v),
\label{eq:nabla-dnfkl}\\
-\nabla_\theta d_n^{\mathrm{RKL}}
&=\sum_{v\in\mathcal{V}}p_S^n(v)
\log r_n(v)\nabla_\theta\log p_S^n(v).
\label{eq:nabla-dnrkl}
\end{align}
The unit baseline in Eq.~\ref{eq:nabla-dnfkl} leaves the gradient unchanged because the score function has zero expectation under the student distribution. Detailed derivations appear in Appendix~\ref{Gradient derivation}. We refer to
\begin{equation}
A_n^{\mathrm{FKL}}(v)=r_n(v)-1,
\qquad
A_n^{\mathrm{RKL}}(v)=\log r_n(v)
\label{eq:implicit-advantages}
\end{equation}
as the implicit token-level advantages.

For any candidate token $v$, the implicit advantages $A_n^{\mathrm{FKL}}(v)$ and $A_n^{\mathrm{RKL}}(v)$ have the same sign: both are positive when the teacher assigns greater probability to the token than the student, negative when it assigns less, and zero when the probabilities coincide. In this representation, the two divergences never disagree on the direction of the update for a given token and differ in the magnitude of the coefficients assigned to each teacher-student discrepancy.

The FKL advantage $r_n(v)-1$ is bounded below by $-1$, limiting the magnitude of negative coefficients for tokens that the student overweights relative to the teacher ($r_n(v)<1$). For underweighted tokens ($r_n(v)>1$), its positive advantage grows linearly without bound. In contrast, the RKL advantage $\log r_n(v)$ is unbounded in both directions: it becomes arbitrarily negative as $r_n(v)$ approaches zero, while its positive values grow only logarithmically.

Since $\log r\leq r-1$ for all $r>0$, with equality only at $r=1$, FKL assigns larger positive advantages when $r_n(v)>1$, whereas RKL assigns more negative advantages when $r_n(v)<1$. Thus, in this advantage representation, divergence selection controls the balance between positive and negative token-level weighting: FKL emphasizes positive feedback for student-underweighted tokens, while RKL emphasizes negative feedback for student-overweighted tokens.

We next analyze the requirements for the properties of the advantage function under different values of the verifier's judgment 
$R(y) \in \{0, 1\}$ for the output $y$ sampled from the student. This motivates the selection of the divergence function according to the verifier's judgment $R(y)$, rather than applying a fixed divergence function for all trajectories as in vanilla OPSD.

\textbf{Correct trajectories ($R(y)=1$).} For a successful trajectory, we aim to reinforce teacher-preferred tokens and absorb the extra probability mass it assigns to valid alternative continuations, without aggressively disrupting an already successful trajectory due to teacher conservatism. This requires a divergence with a gentle, bounded penalty when $r_n(v)<1$ and a strong reward when $r_n(v)>1$. FKL supports this goal through its larger positive advantages and bounded negative advantages. We therefore apply FKL to correct trajectories.

\textbf{Incorrect trajectories ($R(y)=0$).} For an unsuccessful trajectory, we place greater emphasis on suppressing tokens where the student is overconfident relative to the teacher ($r_n(v)<1$), the likely culprits behind the error. The FKL's penalty has a lower bound, limiting the correction speed. However, the unbounded penalty in RKL can achieve a stronger suppression effect. Conversely, for teacher-preferred tokens in a failed rollout, the teacher’s rationalization of $y^\star$ may be locally noisy. Therefore, we tend to adopt a cautious logarithmic damping update rather than an aggressive reinforcement, which is in line with RKL. We thus apply RKL to incorrect trajectories.

This yields an outcome-guided divergence selection rule. When $R(y) = 1$, $d_n = d_n^{\mathrm{FKL}}$, and when $R(y) = 0$, $d_n = d_n^{\mathrm{RKL}}$. It lets the model allocate token-level credit in a manner consistent with the coarse and trajectory-level outcome correctness signal, while retaining the dense and token-level granularity of self-distillation. The analysis of the advantage function $A_{n}(v)$ above is based on a necessary precondition: the guidance signal $p_T^n(v)$ must be reliable, which we examine next.

\subsection{Selective Prefix Distillation}
\label{3.3}

\textbf{Diagnose teacher reliability.}
The teacher is conditioned on student-generated prefixes, which may contain errors or deviate from its preferred reasoning paths. As these prefixes lengthen, the teacher may become less able to produce a correct continuation, a phenomenon known as \emph{prefix failure}. We examine whether this degradation differs between correct and incorrect student trajectories.

We sample responses from Qwen3 models \citep{yang2025qwen3} on OpenThoughts-Math30K \citep{guha2026openthoughts}, DeepScaleR \citep{tan2026deepscaler}, MATH-500 \citep{hendrycksmath2021}, and DeepMath \citep{he2026deepmathk}. We divide the responses into correct and incorrect pools, $\mathcal{D}_c$ and $\mathcal{D}_i$, according to their verifier outcomes. We then concatenate the teacher prompt with student-generated prefixes truncated at varying lengths, and let the teacher resume generation from different token positions across both correct and incorrect student trajectories to complete the remaining reasoning. Teacher completion accuracy (TCA) measures the accuracy rate of these completion results.

Figures~\ref{fig:introb} and~\ref{fig:introd} show a clear difference between the two groups. 
For incorrect student trajectories, the TCA degrades markedly with truncation length. In contrast, for correct trajectories, it remains stable or even improves regardless of truncation position. These observations motivate distillation over the full response when the student succeeds and restrict distillation to a prefix when it fails.

\textbf{Estimate the prefix cutoff.}
We want to seek a boundary separating two regimes of teacher reliability on incorrect trajectories: one where the teacher signal remains valid, and one beyond which guidance quality begins to deteriorate. Following common practice, we use entropy as an indicator of the internal state of the model during token generation. Unlike most work that analyzes token-level entropy \citep{liu2026teacher,jin2026entropyaware}, we argue that single-token entropy is a high-variance and fluctuating signal whose value at any position cannot accurately represent the overall state of the model at that point. We use cumulative average teacher entropy to summarize uncertainty along a prefix and reduce sensitivity to fluctuations at individual positions:
\begin{equation}
\bar{H}_n(y)=\frac{1}{n}\sum_{k=1}^{n}H(p_T^k),
\qquad
H(p)=-\sum_{v\in\mathcal{V}}p(v)\log p(v).
\label{eq:cumulative-entropy}
\end{equation}
For variable-length responses, let $\mathcal{D}_a(n)=\{y\in\mathcal{D}_a:|y|\geq n\}$ contain the trajectories that reach position $n$, where $a\in\{c,i\}$. The outcome-conditioned averages are
\begin{equation}
\bar{H}_n^a
=\frac{1}{|\mathcal{D}_a(n)|}
\sum_{y\in\mathcal{D}_a(n)}\bar{H}_n(y),
\qquad a\in\{c,i\},
\label{eq:outcome-entropy}
\end{equation}
evaluated at positions where both pools are nonempty.

\begin{algorithm}[t]
\caption{Outcome-Guided On-Policy Self-Distillation}
\label{alg:og-opsd}
\begin{algorithmic}[1]
\REQUIRE Training data $\mathcal{S}$; model $\pi_\theta$; verifier $R$; cutoff $N_t$ from Eq.~\ref{nt}
\FOR{each training iteration}
    \STATE Sample a minibatch $\mathcal{B}\subset\mathcal{S}$.
    \FORALL{$(x,y^\star)\in\mathcal{B}$}
        \STATE Generate a complete response $y\sim\pi_\theta(\cdot\mid x)$.
        \STATE Set $N\leftarrow |y|$ and obtain $b\leftarrow R(y)$.
        \IF{$b=1$}
            \STATE $T\leftarrow N$; $\kappa\leftarrow\mathrm{FKL}$.
        \ELSE
            \STATE $T\leftarrow \min\{N,N_t\}$; $\kappa\leftarrow\mathrm{RKL}$.
        \ENDIF
        \STATE Initialize $\ell(x,y^\star)\leftarrow 0$.
        \IF{$T>0$}
            \STATE Compute $p_S^n$ and $p_T^n$ for $n=1,\ldots,T$.
            \STATE $\ell(x,y^\star)\leftarrow T^{-1}\sum_{n=1}^{T}d_n^\kappa$.
        \ENDIF
    \ENDFOR
    \STATE $\mathcal{L}\leftarrow |\mathcal{B}|^{-1}\sum_{(x,y^\star)\in\mathcal{B}}\ell(x,y^\star)$.
    \STATE Update $\theta$ using $\nabla_\theta\mathcal{L}$.
\ENDFOR
\end{algorithmic}
\end{algorithm}
As shown in Figures~\ref{fig:introc} and~\ref{fig:introe}, the two curves almost overlap in the early stage. This indicates that even incorrect reasoning trajectories remain in a regime where the teacher can still provide relatively reliable supervision in early stages, yielding supervision signals of comparable quality to those on correct trajectories. Beyond their last intersection, the two curves gradually diverge, with the incorrect trajectories showing higher cumulative average entropy than the correct trajectories. This divergence is consistent with the hypothesis that accumulating errors cause the trajectory prefix to deviate from the conditional distribution expected by the teacher model, increasing its uncertainty. At this time, the teacher’s supervision signal is no longer consistent with the reliable signal on correct trajectories.  We then consider that the reliability of the teacher signal begins to deteriorate at this point. We use the two curves' last crossing as an empirical cutoff. Let $\Delta_n=\bar{H}_n^i-\bar{H}_n^c$, and let $L$ be the largest position represented in both pools. We define
\begin{equation}
\begin{aligned}
N_t=
\begin{cases}
\max\mathcal{C}, & \mathcal{C}\neq\varnothing,\\
0, & \mathcal{C}=\varnothing.
\end{cases}
\end{aligned}
\label{nt}
\quad \text{where} \quad \mathcal{C}=\left\{n\in\{2,\ldots,L\}:\Delta_{n-1}\Delta_n\leq 0\right\}
\end{equation}
The resulting cutoff $N_t$ is shared across incorrect trajectories.

\textbf{Overall objective.}
Combining outcome-guided divergence selection with prefix restriction gives OG-OPSD. For a response of length $N$, the number of supervised positions $T(y)$ equals $N$ when $R(y)=1$ and equals $\min\{N,N_t\}$ when $R(y)=0$.
The per-response loss is
\begin{equation}
\mathcal{L}_{\mathrm{OG\text{-}OPSD}}(\theta;y)
=\frac{1}{\max\{1,T(y)\}}\sum_{n=1}^{T(y)}
\Bigl[
R(y)d_n^{\mathrm{FKL}}
+\bigl(1-R(y)\bigr)d_n^{\mathrm{RKL}}
\Bigr].
\label{eq:og-opsd-loss}
\end{equation}
Thus, correct responses receive FKL supervision at every position, while incorrect responses receive RKL supervision only within the retained prefix. The truncation applies only when computing the distillation loss.

We average the per-response losses over each minibatch and update the model through the student branch. Algorithm~\ref{alg:og-opsd} summarizes the training procedure.

\section{Experiments}

\subsection{Experimental Setup}
We conduct experiments with the Qwen3-1.7B, Qwen3-4B, and Qwen3-8B Instruct models \citep{yang2025qwen3}. All three models are trained on OpenThoughts-Math30K \citep{guha2026openthoughts}, which contains 29,434 mathematical reasoning examples, each comprising a question, a reference solution, and a final answer. At each distillation iteration, the student model receives only the question and generates its own response, while the teacher model additionally receives the reference solution containing reasoning steps and the final answer. To evaluate the method on multimodal tasks, we also train Qwen3-VL-2B-Instruct \citep{bai2025qwen3} on FinMME \citep{luo2025finmme}. Dataset preparation details are provided in Appendix~\ref{finmme}.

\textbf{Baseline}. 
We compare OG-OPSD with ten baselines for mathematical reasoning. \emph{SFT} \citep{ouyang2022training} trains on reference solutions, while \emph{GRPO} \citep{shao2024deepseekmath} optimizes verifiable rewards based on final-answer correctness. The remaining eight baselines use distillation. \emph{OPSD} \citep{zhao2026selfdistilled} applies a fixed FKL objective uniformly along student-generated trajectories. \emph{EOPD} \citep{jin2026entropyaware} selects the divergence at each token position using teacher entropy. \emph{AVSD} \citep{nguyen2026avsd} combines supervision from multiple privileged teacher views. \emph{REOPOLD} \citep{ko2026scaling} uses the teacher-student log-likelihood ratio as a token-level reward, together with reward clipping and entropy-based sampling. \emph{Purified OPSD} \citep{shen2026purified} uses pointwise mutual information to construct distillation targets that reduce reference-induced interference. \emph{DOPD} \citep{yu2026dopd} dynamically routes teacher-student supervision according to token-level privilege advantage gaps. \emph{ESR} \citep{ziheng2026less} applies a uniform truncation rule to all trajectories. \emph{PW-OPSD} \citep{liu2026teacher} weights token-level distillation losses according to position. For multimodal evaluation, we compare against \emph{SFT} and \emph{SDFT} \citep{shenfeld2026selfdistillation}, a self-distillation method developed for continual learning. We also report the corresponding base model in all comparisons.

\textbf{Benchmarks}. 
We evaluate mathematical reasoning on AIME 2024, AIME 2025, and HMMT 2025, each containing 30 problems, using the official answer keys to determine correctness. Following the OPSD evaluation protocol \citep{zhao2026selfdistilled}, we set the maximum generation length to 38,912 tokens. We generate 12 responses per problem and report the average accuracy over these samples and all problems (Avg@12). We also report the unweighted mean across the three benchmarks.

For multimodal evaluation, we use the FinMME test set and three out-of-distribution (OOD) benchmarks: MathVerse \citep{zhang2024mathverse}, MathVision \citep{wang2024measuring}, and MMMU\textsubscript{Pro} \citep{yue2025mmmu}. To assess cross-domain retention after mathematical post-training, we additionally evaluate the Qwen3 models on LiveCodeBench v5 (LCB v5) \citep{jain2025livecodebench}.

\begin{table*}[t]
\centering
\setlength{\abovecaptionskip}{0pt}
\setlength{\belowcaptionskip}{-10pt}
\caption{Main results on AIME 2024/2025 and HMMT 2025 across Qwen3-1.7B, Qwen3-4B and Qwen3-8B. Scores are Avg@12 and Average is the unweighted mean across benchmarks.}
\resizebox{\linewidth}{!}{
\begin{tabular}{lcccccccccccc}
\toprule
& \multicolumn{4}{c}{Qwen3-1.7B} & \multicolumn{4}{c}{Qwen3-4B} & \multicolumn{4}{c}{Qwen3-8B} \\
\cmidrule(lr){2-5} \cmidrule(lr){6-9} \cmidrule(lr){10-13}
\textbf{Method} & AIME24 & AIME25 & HMMT25 & Avg. & AIME24 & AIME25 & HMMT25 & Avg. & AIME24 & AIME25 & HMMT25 & Avg. \\
\midrule
Base & 51.5 & 36.7 & 23.1 & 37.1 & 74.9 & 66.4 & 42.2 & 61.1 & 75.8 & 65.6 & 43.9 & 61.7 \\
SFT & 48.4 & 36.3 & 22.7 & 35.8 & 70.2 & 62.3 & 43.4 & 58.6 & 72.3 & 64.2 & 42.9 & 59.8 \\
GRPO & 51.1 & 38.3 & 23.7 & 37.7 & 75.6 & 68.1 & 44.4 & 62.7 & 76.4 & 68.9 & 46.7 & 64.0 \\
\midrule
OPSD & 57.2 & 41.1 & 28.8 & 42.3 & 75.6 & 67.9 & 44.5 & 62.6 & 77.8 & 67.5 & 45.8 & 63.7 \\
EOPD & 52.2 & 39.1 & 26.9 & 39.4 & 75.7 & 66.5 & 43.9 & 62.0 & 77.5 & 70.0 & 46.6 & 64.7 \\
AVSD & 57.2 & 38.3 & 24.4 & 39.9 & 75.5 & 68.0 & 44.4 & 62.6 & 75.2 & 69.1 & 46.1 & 63.4 \\
REOPOLD & - & - & - & - & 74.4 & 63.8 & 42.8 & 60.3 & - & - & - & - \\
Purified OPSD & - & - & - & - & 76.3 & 67.2 & \underline{44.7} & 62.7 & - & - & - & - \\
DOPD & - & - & - & - & 74.4 & 66.1 & 41.6 & 60.7 & - & - & - & - \\
ESR & 51.6 & 39.4 & 23.8 & 38.2 & 73.3 & 68.0 & 43.0 & 61.4 & 76.3 & 69.4 & 45.5 & 63.7 \\
PW-OPSD & \underline{57.5} & \underline{41.3} & \underline{31.1} & \underline{43.3} & \underline{76.7} & \underline{68.3} & 44.5 & \underline{63.1} & \underline{78.3} & \underline{70.8} & \underline{47.2} & \underline{65.4} \\
\midrule
\rowcolor{blue!5}
OG-OPSD & \textbf{58.8} & \textbf{41.9} & \textbf{32.5} & \textbf{44.4} & \textbf{77.5} & \textbf{70.2} & \textbf{46.9} & \textbf{64.9} & \textbf{78.9} & \textbf{71.4} & \textbf{48.9} & \textbf{66.4} \\
\bottomrule
\end{tabular}
}
\label{tab:main_math}
\vspace{-10pt}
\end{table*}

\subsection{Main Results}

Table~\ref{tab:main_math} reports the main results on Qwen3-4B. In the experiment, we use the cumulative average teacher entropy to obtain the cutoff $N_t = 100$. OG-OPSD achieves the best performance in all three benchmarks, reaching $77.5$, $70.2$ and $46.9$ on AIME24, AIME25 and HMMT25. OG-OPSD improves over vanilla OPSD by 1.9, 2.3, and 2.4 points on AIME24, AIME25, and HMMT25, respectively, and outperforms the strongest baseline by 1.8 points on average across the three benchmarks. OG-OPSD directly conditions the choice between FKL and RKL on the verifier's correctness signal $R(y)$, which provides a coarser but more reliable anchor than purely distributional heuristics, and no additional hyperparameters or trade-offs are required.

We further investigate how performance gains scale with model size across all three benchmarks. For each model size, our method consistently achieves the maximum performance gain over the evaluated baselines. Moreover, as the model parameter size increases, the advantage gap over vanilla OPSD continues to widen. 
In the HMMT25 dataset, the gap over vanilla OPSD \citep{zhao2026selfdistilled} increases from $2.4$ on Qwen3-4B to $3.1$ on Qwen3-8B, and on AIME25 it grows even more markedly from $2.3$ to $3.9$. Averaged over the three benchmarks, the gain over vanilla OPSD increases steadily with model size, from $2.1$ on Qwen3-1.7B to $2.3$ on Qwen3-4B and $2.7$ on Qwen3-8B. This indicates that the gains from outcome-guided divergence switching and selective truncation are not a phenomenon specific to small models.
A plausible explanation is that larger models possess stronger contextual reasoning capacity and can provide higher-quality teacher supervision signals. 
This reflects the rationality and effectiveness of our dynamic selection of the KL divergence based on the outcome to adapt to the requirements for token reward or penalty magnitudes on different sampling trajectories. 
Due to limitations in computing resources, we are unable to compare all the baselines on the 1.7B and 8B models. However, existing results have already demonstrated to some extent that the observed gain of our method is not limited to the Qwen3-4B.
We also notice that the gain of GRPO is relatively smooth, and even shows a slight negative growth at 1.7B scale on AIME24. This suggests that optimizing solely with sparse final-outcome rewards is challenging for smaller models. By leveraging this binary outcome signal to guide dense distillation supervision, our method remains effective even for small student models.

\subsection{Ablation Study}

\begin{figure}[!ht]
    \centering
    \includegraphics[width=\linewidth]{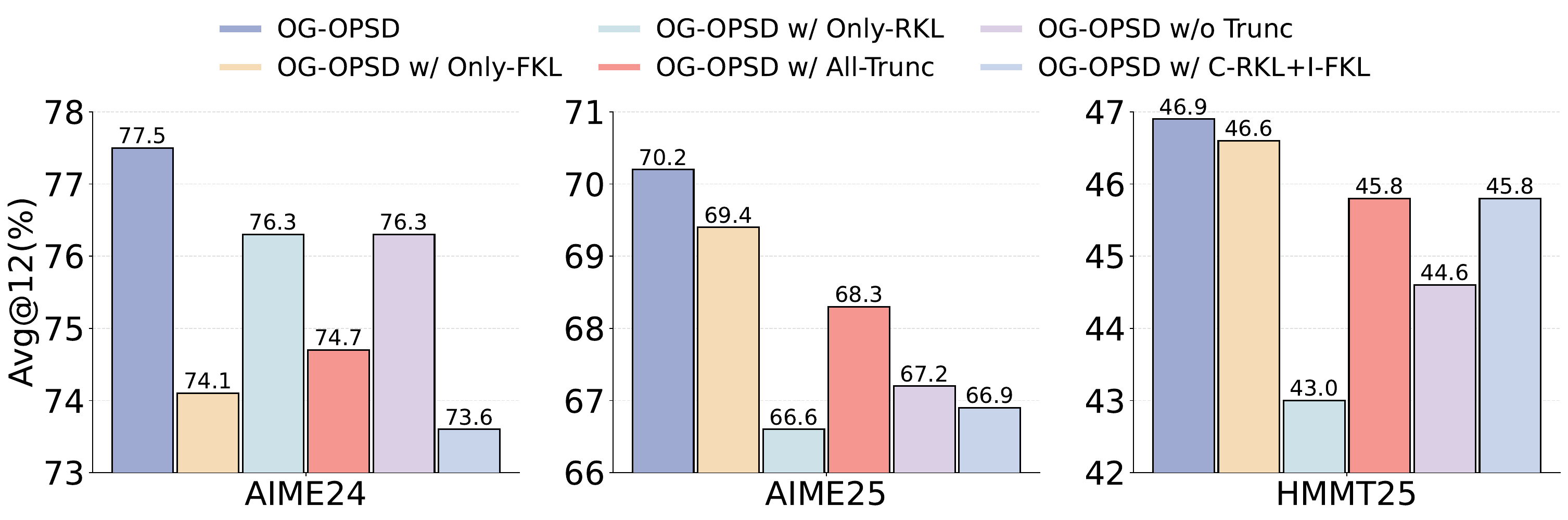}
    \caption{Ablation results of divergence selection strategy and distillation position on AIME24, AIME25 and HMMT25 benchmarks.}
    \label{fig:ablexperiments}
\end{figure}

We conduct ablation studies on both the divergence selection strategy and the distillation position of OG-OPSD. We define five variants: \emph{OG-OPSD w/ Only-FKL} and \emph{OG-OPSD w/ Only-RKL} replace the outcome-guided divergence switching with a fixed FKL and a fixed RKL respectively for all trajectories, while still truncating the incorrect trajectories at position $N_t$. \emph{OG-OPSD w/ All-Trunc} retains our dynamic divergence selection but applies truncation at $N_t$ across all trajectories regardless of correctness. \emph{OG-OPSD w/o Trunc} only uses dynamic divergence selection without any truncation. \emph{OG-OPSD w/ C-RKL+I-FKL} reverses our divergence assignment, using RKL for correct trajectories and FKL for incorrect ones.

\textbf{The selection of the divergence.} As shown in Figure~\ref{fig:ablexperiments}, under identical truncation treatment, OG-OPSD consistently outperforms all alternative divergence-selection variants across the three benchmarks. Both fixing a single divergence (w/ Only-FKL, w/ Only-RKL) and reversing the divergence assignment (w/ C-RKL+I-FKL) lead to clear performance drops. This provides evidence that the gains of OG-OPSD do not simply come from having some adaptive mechanism, but specifically from correctly pairing FKL with correct trajectories and RKL with incorrect ones, consistent with our analysis in Section~\ref{3.2}.

\textbf{The selection of the distillation position.} Replacing outcome-guided truncation with all truncation (w/ All-Trunc) also yields consistent performance drops across all benchmarks, showing that applying cutoff to all trajectories regardless of their correctness, such as ESR \citep{ziheng2026less}, discards the valid supervision signal from the correct trajectories. 
\begin{wrapfigure}{r}{0.46\linewidth}
    \setlength{\abovecaptionskip}{10pt}
    \setlength{\belowcaptionskip}{-10pt}
    
    \centering
    \includegraphics[width=\linewidth]{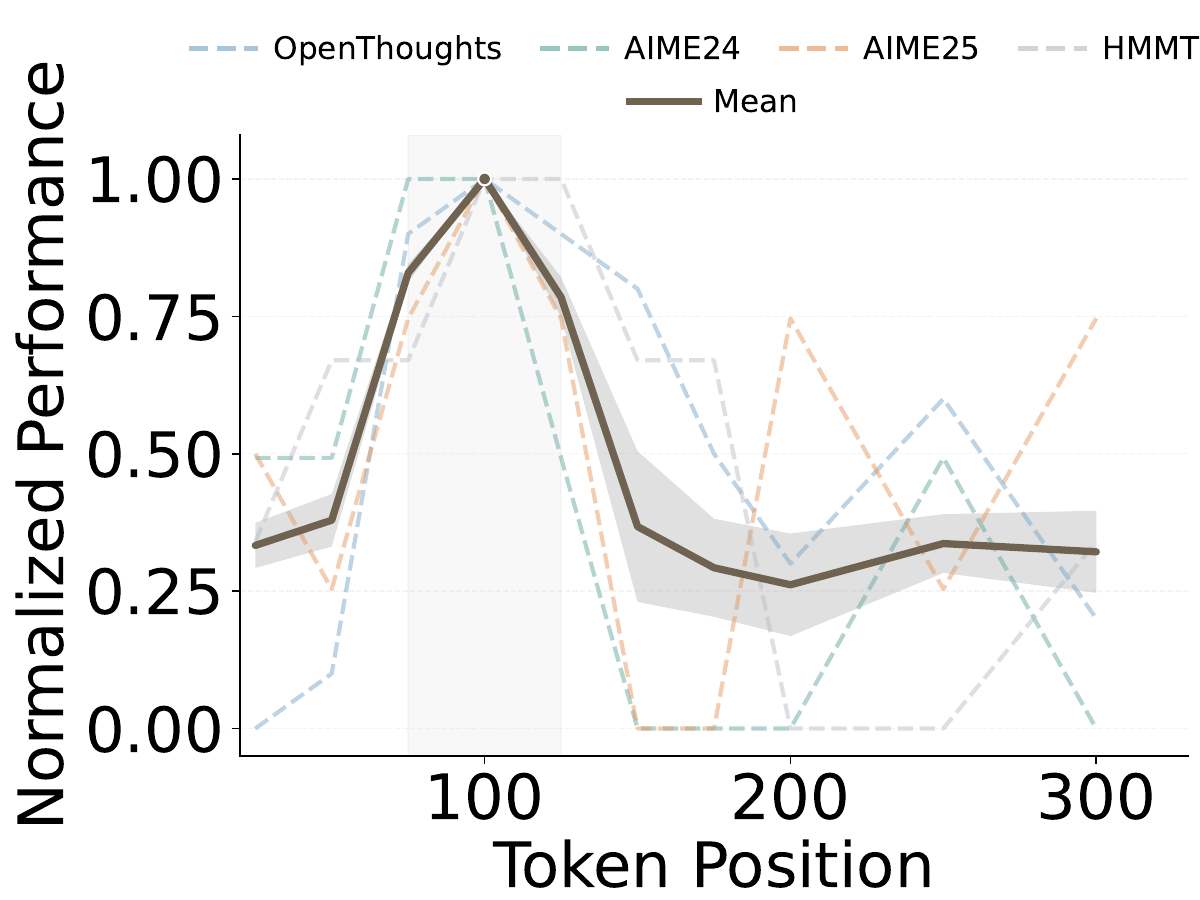}
    \vspace{-2em}
    \caption{Train with different $N_t$ and evaluate on multiple datasets.}
    \label{fig:abl}
    \vspace{-10pt}
\end{wrapfigure}
The decline in the performance of \emph{OG-OPSD w/o Trunc} indicates the necessity of considering the reliability of the teacher's guidance signal. Before training, using Eq.~\ref{nt}, we calculate the truncation parameter $N_t$ as $100$ on the training data. We further examine the sensitivity of $N_t$ in Figure~\ref{fig:abl}. We use different values of $N_t$ to train Qwen3-4B. After calculating Pass@1 on OpenThoughts, AIME24, AIME25, and HMMT25, we normalize the results and also take the average. The performance of the model peaks at $N_t=100$ and remains relatively stable within the adjacent range. This indicates that $N_t$, derived from the cumulative average teacher entropy, is robust to moderate perturbation and does not require precise tuning to be effective.
\subsection{Multimodal and OOD results}
\begin{table}[t]
\centering
\scriptsize
\begin{minipage}[t]{0.49\textwidth}
\centering
\setlength{\abovecaptionskip}{0pt}
\setlength{\belowcaptionskip}{-10pt}
\caption{Evaluation on FinMME and OOD datasets after training on FinMME.}
\begin{tabular}{lccc>{\columncolor{blue!5}}c}
\toprule
 & Base & SFT & SDFT & OG-OPSD \\
\midrule
FinMME & 35.0 & 32.2(\textcolor{red}{$\downarrow$ 2.8}) & 37.2(\textcolor{green}{$\uparrow$ 2.2}) & 38.2(\textcolor{green}{$\uparrow$ 3.2}) \\
\midrule
MathVerse & 31.2 & 24.5(\textcolor{red}{$\downarrow$ 6.7}) & 30.3(\textcolor{red}{$\downarrow$ 0.9}) & 32.4(\textcolor{green}{$\uparrow$ 1.2}) \\
MathVision & 14.8 & 14.5(\textcolor{red}{$\downarrow$ 0.3}) & 12.8(\textcolor{red}{$\downarrow$ 2.0}) & 17.4(\textcolor{green}{$\uparrow$ 2.6}) \\
MMMU\textsubscript{Pro} & 33.7 & 32.5(\textcolor{red}{$\downarrow$ 1.2}) & 33.5(\textcolor{red}{$\downarrow$ 0.2}) & 34.4(\textcolor{green}{$\uparrow$ 0.7}) \\
\bottomrule
\end{tabular}
\label{tab:finmme_result}
\end{minipage}
\hfill
\begin{minipage}[t]{0.49\textwidth}
\centering
\setlength{\abovecaptionskip}{0pt}
\setlength{\belowcaptionskip}{-10pt}
\caption{Evaluation on LCB v5 after training on OpenThoughts-Math30K.}
\begin{tabular}{lccc}
\toprule
Method & Qwen3-1.7B & Qwen3-4B & Qwen3-8B \\
\midrule
Base & 32.8 & 60.3 & 53.5 \\
\midrule
OPSD & 18.9(\textcolor{red}{$\downarrow$ 13.9}) & 49.2(\textcolor{red}{$\downarrow$ 11.1}) & 45.5(\textcolor{red}{$\downarrow$ 8.0}) \\
PW-OPSD & 17.7(\textcolor{red}{$\downarrow$ 15.1}) & 46.2(\textcolor{red}{$\downarrow$ 14.1}) & 47.8(\textcolor{red}{$\downarrow$ 5.7}) \\
\rowcolor{blue!5}
OG-OPSD & 24.6(\textcolor{red}{$\downarrow$ 8.2}) & 54.1(\textcolor{red}{$\downarrow$ 6.2}) & 52.5(\textcolor{red}{$\downarrow$ 1.0}) \\
\bottomrule
\end{tabular}
\label{tab:lcb_v5}
\end{minipage}
\vspace{-10pt} 
\end{table}

Table~\ref{tab:finmme_result} reports the results after training Qwen3-VL-2B-Instruct on FinMME. OG-OPSD achieves the highest scores on both FinMME and the three OOD benchmarks among the evaluated methods. On FinMME, it scores $38.2$, improving over the base model by $3.2$ points. On MathVerse, MathVision, and MMMU\textsubscript{Pro}, it exceeds SDFT by $2.1$, $4.6$, and $0.9$ points, respectively. It also improves over the base model on all three OOD benchmarks, demonstrating its applicability to multimodal tasks and its benefits beyond the training domain.

Table~\ref{tab:lcb_v5} evaluates coding performance after mathematical post-training on OpenThoughts-Math30K. All three distillation methods reduce LiveCodeBench v5 scores relative to the base models, but OG-OPSD consistently incurs the smallest decline. Its score reductions are $8.2$, $6.2$, and $1.0$ points at the 1.7B, 4B, and 8B scales, compared with $13.9$, $11.1$, and $8.0$ points for vanilla OPSD. These results indicate that OG-OPSD better preserves coding performance while improving mathematical reasoning. This might be because we truncate the incorrect trajectories by using cumulative average entropy, which may reduce interference with the model’s existing knowledge caused by the low-quality guidance from the teacher in the later stages of the incorrect trajectories.

\section{Conclusion}
We present OG-OPSD, which uses binary outcome rewards to guide divergence selection and determine which token positions receive supervision. Our analysis shows that the implicit token-level advantages of FKL and RKL have matching signs but different magnitudes, motivating FKL for correct trajectories and RKL for incorrect ones. We further identify an outcome-dependent pattern of prefix failure: teacher continuation accuracy declines with prefix length on incorrect trajectories but remains stable or improves on correct ones. Accordingly, OG-OPSD distills correct trajectories in full and restricts distillation on incorrect trajectories to prefixes whose length is estimated from cumulative average teacher entropy.
Experiments on Qwen3 and Qwen3-VL show consistent improvements over the evaluated baselines, with ablations supporting the contributions of both divergence selection and selective truncation. These findings highlight how coarse but reliable outcome signals can guide the use of dense self-distillation supervision. Future work could extend this approach to graded rewards and per-trajectory adaptive truncation.

\bibliography{iclr2027_conference}
\bibliographystyle{iclr2027_conference}

\newpage

\appendix
\section{Gradient Derivation}
\label{Gradient derivation}

In this section we derive Eq.~\ref{eq:nabla-dnfkl} and Eq.~\ref{eq:nabla-dnrkl}. For analytical clarity, we first derive the gradients of the unclipped KL objectives. In implementation, we additionally clip individual KL components as described in Appendix~\ref{finmme}.

\paragraph{Notation and assumptions.}
Fix a position $n$ in a student-generated rollout, so that the prefix $s_n$ is a given constant. For brevity, write
\begin{equation}
    p_i \triangleq p^n_S(v_i ), \qquad q_i \triangleq p^n_T(v_i), \qquad r_i \triangleq r_n(v_i) = \frac{q_i}{p_i}.
\end{equation}
Both are categorical distributions over $\mathcal{V}$, $p_i, q_i > 0$ and $\sum_{i=1}^{|\mathcal{V}|} p_i = \sum_{i=1}^{|\mathcal{V}|} q_i = 1$. The teacher distribution is treated as a fixed target:
\begin{equation}
    \nabla_\theta q_i = 0, \qquad \nabla_\theta p_i = p_i \nabla_\theta \log p_i .
\end{equation}
The second equality is the log-derivative trick.

Since $\sum_i^{|\mathcal{V}|} p_i = 1$ holds for all $\theta$,
\begin{equation}
    \sum_{i}^{|\mathcal{V}|} p_i \nabla_\theta \log p_i = \sum_{i}^{|\mathcal{V}|} \nabla_\theta p_i = \nabla_\theta \sum_{i}^{|\mathcal{V}|} p_i = \nabla_\theta 1 = 0.
    \label{eq:score-zero}
\end{equation}
Consequently, for any constant $c$ that does not depend on $i$,
\begin{equation}
    \sum_{i}^{|\mathcal{V}|} p_i \, c \, \nabla_\theta \log p_i = 0 .
    \label{eq:baseline}
\end{equation}

\subsection{Forward KL}

According to the definition of Forward KL divergence,
\begin{equation}
    d_n^{\mathrm{FKL}} = D_{\mathrm{KL}}(p_T^n \Vert p_S^n) = \sum_{i}^{|\mathcal{V}|} q_i \left( \log q_i - \log p_i \right).
\end{equation}
The $q_i$ is a constant with respect to $\theta$, so the term $q_i \log q_i$ vanishes under differentiation.
\begin{align}
    \nabla_\theta d_n^{\mathrm{FKL}} &= \nabla_\theta \sum_{i}^{|\mathcal{V}|} q_i \left( \log q_i - \log p_i \right) \\
    &=\sum_{i}^{|\mathcal{V}|} \left( \nabla_\theta \left(q_i \log q_i\right)- \nabla_\theta \left(q_i \log p_i\right) \right) \\
    &=
    -\sum_{i}^{|\mathcal{V}|} q_i \nabla_\theta \log p_i.
\end{align}
To write this in the form $\sum_i p_i A(v_i) \nabla_\theta \log p_i$, we use $q_i = p_i r_i$:
\begin{equation}
    -\nabla_\theta d_n^{\mathrm{FKL}} = \sum_{i}^{|\mathcal{V}|} p_i \, r_i \, \nabla_\theta \log p_i .
\end{equation}
Subtracting the zero term in Eq.~\ref{eq:baseline} with $c = 1$:
\begin{align}
    -\nabla_\theta d_n^{\mathrm{FKL}}
    &= \sum_{i}^{|\mathcal{V}|} p_i \, r_i \, \nabla_\theta \log p_i - \sum_{i}^{|\mathcal{V}|} p_i \, \nabla_\theta \log p_i \\
    &= \sum_{i}^{|\mathcal{V}|} p_i \left( r_i - 1 \right) \nabla_\theta \log p_i ,
\end{align}
which is exactly Eq.~\ref{eq:nabla-dnfkl}, with advantage $A_n^{\mathrm{FKL}}(v) = r_n(v) - 1$. \hfill$\square$

\subsection{Reverse KL}

According to the definition of Reverse KL divergence,
\begin{equation}
    d_n^{\mathrm{RKL}} = D_{\mathrm{KL}}(p_S^n \Vert p_T^n) = \sum_{i}^{|\mathcal{V}|} p_i \left( \log p_i - \log q_i \right).
\end{equation}
Now $p_i$ appears both as the weight and inside the logarithm, so we apply the product rule:
\begin{equation}
    \nabla_\theta d_n^{\mathrm{RKL}}
    = \sum_{i}^{|\mathcal{V}|} \nabla_\theta p_i \left( \log p_i - \log q_i \right)
    + \sum_{i}^{|\mathcal{V}|} p_i \, \nabla_\theta \left( \log p_i - \log q_i \right).
\end{equation}
For the second sum, $\nabla_\theta \log q_i = 0$ and $\sum_i^{|\mathcal{V}|} p_i \nabla_\theta \log p_i = 0$ by Eq.~\ref{eq:score-zero}, so it's equal to zero.

For the first sum, we apply the log-derivative trick:
\begin{align}
    \sum_{i}^{|\mathcal{V}|} \nabla_\theta p_i \left( \log p_i - \log q_i \right)
    &= \sum_{i}^{|\mathcal{V}|} p_i \left( \log p_i - \log q_i \right) \nabla_\theta \log p_i \\
    &=\sum_{i}^{|\mathcal{V}|} p_i \left( -\log r_i \right) \nabla_\theta \log p_i .
\end{align}
Combining the two,
\begin{equation}
    -\nabla_\theta d_n^{\mathrm{RKL}} = \sum_{i}^{|\mathcal{V}|} p_i \log r_i \, \nabla_\theta \log p_i ,
\end{equation}
which is exactly Eq.~\ref{eq:nabla-dnrkl}, with advantage $A_n^{\mathrm{RKL}}(v) = \log r_n(v)$. \hfill$\square$

\section{Implementation Details}
\label{finmme}
\textbf{Details of FinMME Dataset Processing.} FinMME \citep{luo2025finmme} is a comprehensive benchmark dataset designed for evaluating multimodal large language models in the financial domain. It comprises approximately 11,000 high-quality financial samples covering 18 subfields of finance and six major asset classes. In terms of question types, the dataset includes single-choice questions, multiple-choice questions, and numerical answer questions.
In this work, we first filter out the samples with excessively high image resolution. Then, we adjust the prompts for different question types and extracted subsets. The sampling process ensures that options A, B, C, and D appear equally frequently in all answers. Finally, a training set of 1200 samples is constructed, including 600 multiple-choice questions and 600 numerical questions. A test set of 600 samples is also obtained, consisting of 300 multiple-choice questions and 300 numerical questions. We use the Qwen3-VL-32B-Instruct to generate reference solutions for this dataset as privileged information for the teacher model.

\textbf{Training.} Our training setup follows vanilla OPSD \citep{zhao2026selfdistilled}. All attention and multilayer perceptron projections are adapted using LoRA with rank $64$ and scale $128$. The privileged-teacher forward pass is evaluated with the LoRA adapter disabled, so gradient updates are applied only to the student's LoRA parameters. We optimize with a learning rate of $5 \times 10^{-6}$, gradient-norm clipping at $0.1$, bfloat16 mixed precision, and gradient checkpointing. 
The student samples one on-policy rollout per problem at temperature $1.1$, top-p $0.95$, top-k $20$, generating up to $1,024$ new tokens and the training context length is capped at $20,000$ tokens. During rollout generation, the student receives only the problem statement, rendered with the Qwen3 chat template and \texttt{enable\_thinking=False}. The teacher receives the problem together with the reference solution (including the final answer). The specific prompts are presented in Table~\ref{tab:template}.
We use a per-device batch size of $4$ with $2$ gradient accumulation steps, and the main experiments utilize $4$ NVIDIA RTX Pro 6000 GPUs. The data collator right-pads prompts within a batch. Every directed loss term is computed over the full vocabulary at each non-padding completion token. Both teacher and student logits are temperature-scaled by $1.1$ prior to computing the KL divergence, and, consistent with vanilla OPSD \citep{zhao2026selfdistilled}, individual KL components are clipped with a cap of $c=0.05$. All baselines use their published default hyperparameters.

\begin{table}[h]
\centering
\begin{tabular}{l p{7cm}}
\toprule
View & User-message template \\
\midrule
\textbf{Student} & \textbf{Problem:} \{problem\} \\
& Please reason step by step, and put your final answer within \textbackslash boxed\{\}. \\[4pt]
\textbf{Teacher} & \textbf{Problem:} \{problem\} \\
& Here is a reference solution to this problem: \\
& === Reference Solution Begin === \\
& \{reference completion\} \\
& === Reference Solution End === \\
& After reading the reference solution above, make sure you truly understand the reasoning behind each step --- do not copy or paraphrase it. Now, using your own words and independent reasoning, derive the same final answer to the problem above. Think step by step, explore different approaches, and don't be afraid to backtrack or reconsider if something doesn't work out: \\
& Please reason step by step, and put your final answer within \textbackslash boxed\{\}. \\
\bottomrule
\end{tabular}
\caption{User-message templates for the \textit{Student} view without privileged information and the reference-conditioned \textit{Teacher} view.}
\label{tab:template}
\end{table}

\textbf{Evaluation.}
We use a maximum generation length of $38,912$ tokens: $\texttt{max\_new\_tokens}=38912$, $M=12$ samples per problem, temperature $1.0$, top-p $0.95$, top-k disabled, and \texttt{enable\_thinking=True}, matching the vanilla OPSD evaluation setup. For a single problem with $M=12$ generated solutions $\{ y^{(1)}, \ldots, y^{(M)} \}$, the predicted answer is extracted as the content of the last \texttt{\textbackslash boxed\{...\}} in $y^{(i)}$. Let $c_i \in \{0, 1\}$ indicate whether $y^{(i)}$ is graded correct against the gold answer. We report the Avg@12 on AIME24, AIME25, and HMMT25.
\begin{equation}
    \text{Avg@} 12 = \frac{1}{12} \sum_{i=1}^{12} c_i.
\end{equation}

\section{More Results On Qwen3-1.7B}
We observe the same trend on Qwen3-1.7B as the 4B and 8B scale models in the main text. The results of TCA are averaged over OpenThoughts-Math30K, DeepScaleR, MATH-500, and DeepMath datasets consistent with Figures~\ref{fig:introb} and~\ref{fig:introd}.
\begin{figure}[b]
    \centering
    \setlength{\abovecaptionskip}{0pt}
    \setlength{\belowcaptionskip}{0pt}
    \begin{subfigure}[c]{0.38\linewidth} 
        \centering
        \makebox[\linewidth][c]{\hspace{-1mm}\includegraphics[width=0.92\linewidth]{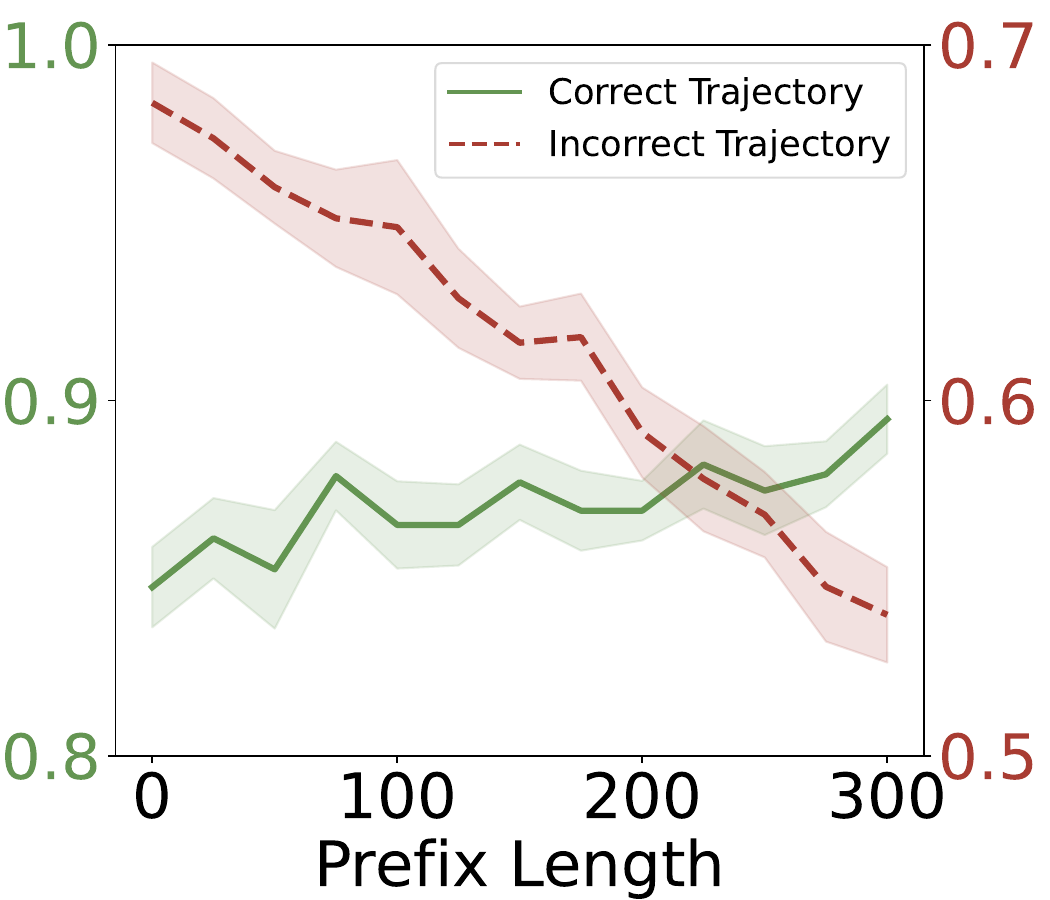}}
        \caption{Teacher completion accuracy (TCA)}
        \label{tca1b}
    \end{subfigure}
    \begin{subfigure}[c]{0.38\linewidth}
        \centering
        \makebox[\linewidth][c]{\hspace{1mm}\includegraphics[width=0.92\linewidth]{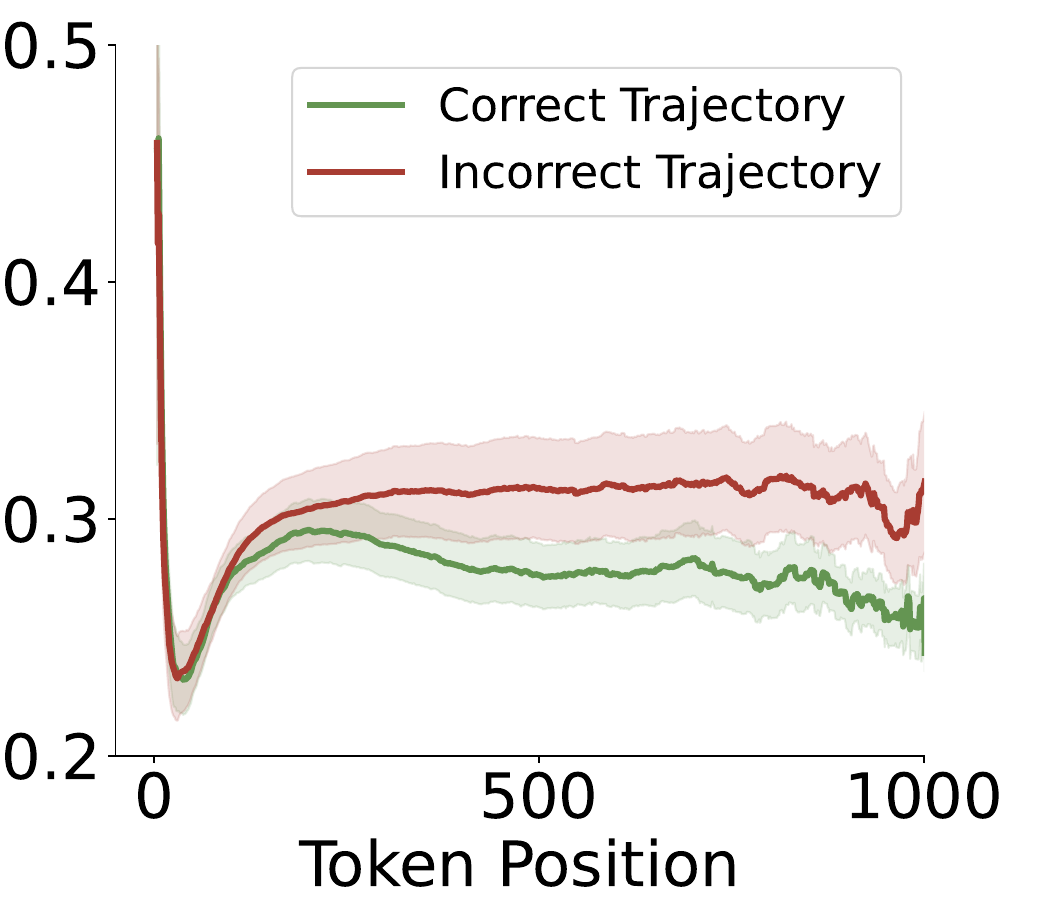}}
        \caption{Cumulative average teacher entropy}
        \label{tcae1b}
    \end{subfigure}
    \caption{TCA and cumulative average teacher entropy on Qwen3-1.7B.}
    \label{qwen1b}
    \vspace{0pt}
\end{figure}

\end{document}